# A Multi-Level Deep Ensemble Model for Skin Lesion Classification in Dermoscopy Images

Yutong Xie, Jianpeng Zhang, and Yong Xia*

National Engineering Laboratory for Integrated Aero-Space-Ground-Ocean Big Data Application Technology, School of Computer Science and Engineering, Northwestern Polytechnical University, Xi'an, PR China, 710072

Corresponding Author's Email: yxia@nwpu.edu.cn

**Abstract.** A multi-level deep ensemble (MLDE) model that can be trained in an 'end to end' manner is proposed for skin lesion classification in dermoscopy images. In this model, four pre-trained ResNet-50 networks are used to characterize the multiscale information of skin lesions and are combined by using an adaptive weighting scheme that can be learned during the error back propagation. The proposed MLDE model achieved an average AUC value of 86.5% on the ISIC-skin 2018 official validation dataset, which is substantially higher than the average AUC values achieved by each of four ResNet-50 networks.

## 1 Introduction

Skin cancer is one of the most common forms of cancers in the United States and many other countries, with 5 million cases occurring annually [1]. Dermoscopy images [2] is one of the essential means to improve diagnostic performance and reduce melanoma deaths [3]. Automated classification of skin lesions, particularly the melanoma, in dermoscopy images has drawn a lot of research attention, since the manual classification requires a lot of skills and concentration, and is expensive and prone to operator-related bias. Among the skin lesion classification methods proposed in the literature, deep convolutional neural network (DCNN) based solutions [4] have shown improved performance [5]. However, this classification task remains a challenge largely due to the insufficiency of training data, which relates to the work required in acquiring the dermoscopy images and then in annotation [6]. Thus, it is difficult for DCNNs to achieve the same success on skin lesion classification, which has only thousands of data, as they have done in the ImageNet Challenge [7], which has tens of millions of data. To solve this "small-data" learning problem, in this paper we propose a multi-level deep ensemble (MLDE) model for skin lesion classification in dermoscopy images. This model consists of four pre-trained ResNet-50 networks, which are fed with multiscale region of interests (ROIs) extracted in dermoscopy images. The probability scores of produced by these four networks are fused with an adaptive weighting scheme that can be learned during the error back propagation. The proposed model has been evaluated on the ISIC-skin 2018 official validation dataset [3, 8] and achieved an online validation score of 90.2%.

## 2 Dataset

The ISIC-skin 2018 challenge dataset [3, 8] was used for this study. This dataset is the largest publicly available one, consisting of 10015 training, 193 validation, and 1512 test images screened for both privacy and quality assurance. Lesions in dermoscopy images are all paired with a gold standard (definitive) diagnosis, i.e. melanoma, melanocytic nevus, basal cell carcinoma, actinic keratosis, benign keratosis, dermatofibroma and vascular lesion.

## 3 Method

The proposed MLDE model is composed of four pre-trained ResNet-50 networks, whose learned probability scores are fused by using an adaptive weighting scheme in the final classification layer (see Fig. 1). In this study, seven MLDE models were constructed for seven binary classification tasks, respectively.

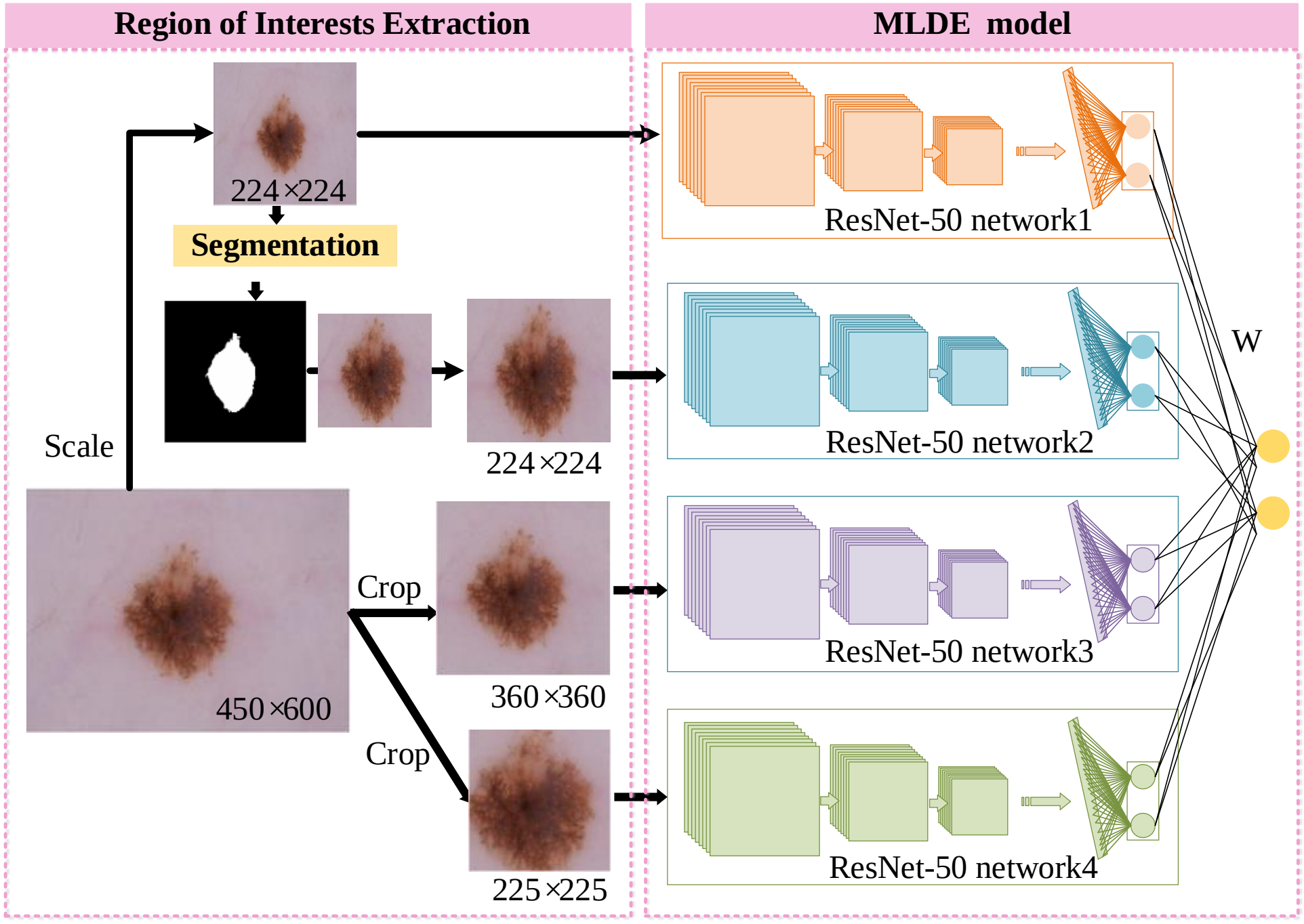

**Fig. 1.** Framework of our proposed MLDE algorithm

### 3.1 Transferable DCNN

A ResNet-50 network contains 50 learnable layers, including consequently a 7×7 convolutional layer that produces 64 feature maps, a 3×3 max pooling layer, four bottleneck blocks, an average pooling layer, and a fully connected layer with 1000 neurons. Each bottleneck block consists of three convolutional layers of size 1×1, 3×3, and 1×1, respectively. From the first to the fourth bottleneck block, the number of feature maps (channels) increases, whereas the size of feature maps decreases.

To transfer the image representation ability learned on large scale image databases to skin lesion characterization, the parameters used to initialize each ResNet-50 network have been converged by training on the ImageNet dataset [7, 9]. To adapt ResNet-50 to skin lesion classification, its last fully connected layer was is replaced by a fully connected layer with two neurons, which are randomly initialized by using the Xaiver algorithm. Each modified ResNet-50 network is then fine-tuned in a layer-wise manner, starting with tuning only the last layer.

### 3.2 Multi-Level Deep Networks and Model Ensemble

All training images are resized to 224×224 by using the bilinear interpolation, and then are used to fine tune the first ResNet-50 network. Since the skin lesion occupies only a small part of an image, shrinking the image may lead the lesion to become too small to be classified. To address this issue, other three

ResNet-50 networks are constructed to use the multiscale information surrounding skin lesions in three steps. First, we design a segmentation network, whose architecture is shown in Fig. 2, to locate each skin lesion. Next, we identify a rectangular ROI that encapsulates the skin lesion and reduces the interference of surrounding background, and then resized the ROI to 224×224 using the bilinear interpolation to fine tune the second ResNet-50. Finally, based on the lesion segmentation, we crop two rectangular image patches of size 360×360 and 225×225 from the original training image, which are centered on the centroid of the lesion, and use these patches to fine tune the third and fourth ResNet-50 networks, respectively.

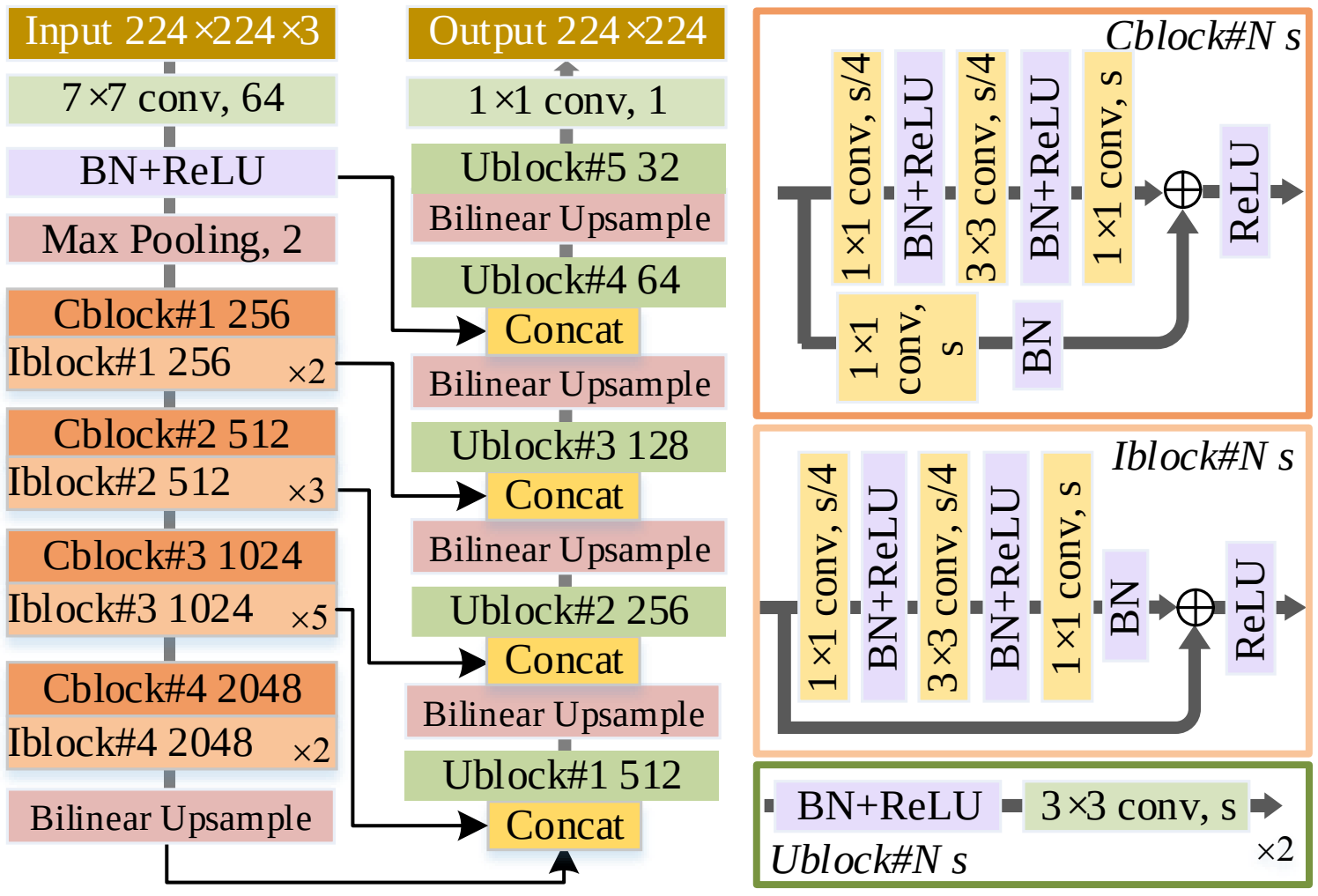

**Fig. 2.** Architecture of the network used for skin lesion segmentation.

The probability scores produced by these four ResNet-50 networks are concatenated to form a new classification layer, which consists of a fully-connected layer with two neuron followed by the softmax function. The output of the classification layer is the final prediction of the proposed MLDE model. Specifically, for the $n$-th input image, we denoted the corresponding output of $k$-th ResNet-50 by $M_{nk} \in R^2$. Then, the output of the MLDE model can be formulated as

$$P_n = f\left(\sum_{k=1}^{4}\sum_{j=1}^{2} W_{kj} M_{nkj}\right) \quad (1)$$

where $\left\{W_{kj}: k = 1,2,\cdots,4; j \in \{1,2\}\right\}$ is the assemble weights between the output layer of each ResNet-50 and the final output neurons. The assemble weights can be updated during the error back propagation process, and hence the MLDE model can be trained in an 'end to end' way.

To train the proposed MLDE model, we chose the min-batch stochastic gradient descent (SGD) with a batch size of 32 as the optimizer and set the learning rate to 0.001 and maximum epoch to 500. Moreover, we randomly selected 10% of training patches to form a validation set and terminated the training process if the error on the validation set stops decreasing but the error on the other training patches continues to decline.

## 4 Results

We applied the MLDE model to the ISIC-skin 2018 official validation dataset. Since the ground truth are not available to the public, all results were processed by the online submission system. Participants were ranked by the mean value of the seven classifiers' area under the receiver operating characteristic curves (AUCs). Table 1 shows the results obtained by using each of four individual ResNet-50 networks and the proposed MLDE model. It reveals that our MLDE model achieved an average AUC value of 90.2%, substantially higher than the average AUC values achieved by each individual ResNet-50 network.

**Table 1.** Comparison of the skin lesion classification for different methods.

| Methods | MLDE model | ResNet-50 network1 | ResNet-50 network2 | ResNet-50 network3 | ResNet-50 network4 |
|---|---|---|---|---|---|
| Average AUC (%) | 86.5 | 83.1 | 85.2 | 84.1 | 83.0 |

## 5 Conclusion

In this paper, we propose the MLDE model for skin lesion classification in dermoscopy images. This model jointly uses four pre-trained ResNet-50 networks to characterize the multiscale information of skin lesions. Our results show that the proposed MLDE model achieved a substantially higher average AUC value on the ISIC-skin 2018 official validation dataset than each of four ResNet-50 networks.

**Acknowledgements.** This work was supported in part by the National Natural Science Foundation of China under Grants 61771397 and 61471297. We acknowledge the efforts devoted by the International Skin Imaging Collaboration (ISIC) to collect and share the skin lesion classification dataset.